\documentclass[conference,letter]{IEEEtran}
\ifCLASSINFOpdf
   \usepackage[pdftex]{graphicx}
   \graphicspath{{pics/}{../pdf/}{../jpeg/}}
   \DeclareGraphicsExtensions{.pdf,.jpeg,.png}
\else
   \usepackage[dvips]{graphicx}
   \graphicspath{{../eps/}}
   \DeclareGraphicsExtensions{.eps}
\fi
\usepackage{amsmath}
\usepackage{amsfonts}
\usepackage{ragged2e}

\DeclareMathOperator*{\argmin}{arg\,min}

\usepackage{array}
\ifCLASSOPTIONcompsoc
  \usepackage[caption=false,font=normalsize,labelfont=sf,textfont=sf]{subfig}
\else
  \usepackage[caption=false,font=footnotesize]{subfig}
\fi
\usepackage{url}
\usepackage{color}

\IEEEoverridecommandlockouts
\IEEEpubid{\makebox[\columnwidth]{978-1-5386-3159-1/17/\textdollar31.00 \copyright2017 IEEE \hfill} \hspace{\columnsep}\makebox[\columnwidth]{ }}

\begin{document}
\title{Adaptive Smoothing in fMRI Data \\Processing Neural Networks}

\author{\IEEEauthorblockN{Albert Vilamala}
\IEEEauthorblockA{Technical University of Denmark\\alvmu@dtu.dk}\vspace{-4.5ex}
\and
\IEEEauthorblockN{Kristoffer Hougaard Madsen}
\IEEEauthorblockA{Danish Research Centre for Magnetic Resonance\\kristofferm@drcmr.dk}\vspace{-4.5ex}
\and
\IEEEauthorblockN{Lars Kai Hansen}
\IEEEauthorblockA{Technical University of Denmark\\lkai@dtu.dk}\vspace{-4.5ex}
}

\maketitle
\begin{abstract}
Functional Magnetic Resonance Imaging (fMRI) relies on multi-step data processing pipelines to accurately determine brain activity; among them, the crucial step of spatial smoothing. These pipelines are commonly suboptimal, given the local optimisation strategy they use, treating each step in isolation. With the advent of new tools for deep learning, recent work has proposed to turn these pipelines into end-to-end learning networks. This change of paradigm offers new avenues to improvement as it allows for a global optimisation. The current work aims at benefitting from this paradigm shift by defining a smoothing step as a layer in these networks able to adaptively modulate the degree of smoothing required by each brain volume to better accomplish a given data analysis task. The viability is evaluated on real fMRI data where subjects did alternate between left and right finger tapping tasks.
\end{abstract}


% \fi
\IEEEpeerreviewmaketitle

%%%%%%% INTRODUCTION %%%%%%%%
\section{Introduction}
The use of non-invasive functional Magnetic Resonance Imaging (fMRI) techniques for determining brain activity requires a set of data processing steps that transforms raw data into validated elements suitable for statistical analysis.

%\begin{figure}[htb!]
%\centering
%\includegraphics[width=0.4\textwidth]{fmri_pipeline}
%\caption{fMRI data processing pipeline.}
%\label{fig:fmri_pipeline}
%\end{figure}

%One crucial preprocessing step often applied in practise is the \emph{spatial smoothing}, usually consisting in sliding a Gaussian-shaped filter over brain volumes with the purpose of locally averaging voxels' intensities. According to the literature \cite{Lindquist2008}, there are three main reasons that make this step sensible:
%\begin{enumerate}
%\item Ensure %the fulfilment of 
%Random Field Theory (RFT) assumptions in posterior steps of the pipeline, such as correcting for multiple comparisons when applying the General Linear Model (GLM) technique to locate brain activity.
%\item Reduce random noise in the voxels, hence increasing Signal-to-Noise Ratio (SNR) that leads to an improvement of the statistical techniques to detect true activations.
%\item Account for intersubject anatomical differences, which might still exist despite affine and non-linear transformations in the spatial normalisation step of the pipeline.
%\end{enumerate}
Gaussian filter to average local voxel intensities is an important preprocessing step. Spatial smoothing serves several purposes \cite{Lindquist2008}: local averaging reduces uncorrelated random noise in the voxel, hence increasing the Signal-to-Noise Ratio (SNR) leading to improved statistical power to detect true functional brain activation; also, spatial smoothing serves to eliminate unimportant anatomical details across subjects, that are preserved despite affine and non-linear transformations, which are common spatial normalisation steps in the pipeline; additionally, smoothing can ensure that assumptions typically made to enable multiple comparison correction using Random Field Theory (RFT) for locating brain activation are fulfilled.

Notice that the resolution of each brain volume is decreased by applying the smoothing step, meaning that an appropriate trade-off between the original volume and the degree of smoothing to be applied is sought. This compromise is governed by a single parameter stating the width of the Gaussian filter. In spite of the importance of this parameter, there is no established method to automatically select its most appropriate value for every situation, being often set according to best practices or relying on each scientist's expertise. %relying on the expertise of the scientists carrying out the experiments.

Previous attempts to build a system capable of adaptively smoothing brain volumes in fMRI include \cite{Penny2005}, which defines a Bayesian approach for the General Linear Model (GLM) able to determine the optimal amount of smoothing for each regressor; \cite{Tabelow2006} and \cite{Almodovar2017} which smooth the resulting statistical parametric maps in a GLM instead of the input volumes, while providing an appropriate activation threshold based on the spatial correlation; and  \cite{Yue2010}, that uses non-stationary spatial Gaussian Markov random fields, allowing the spatial extent of smoothing to vary across both space and time.

A recent work \cite{Vilamala2016} proposes a change of paradigm by converting fMRI data processing pipelines into deep neural networks with the purpose of optimising the pipelines end-to-end, instead of current state of the art, which optimises them locally at each step of the pipeline. In the same study, authors propose a generalisation of the Spatial Transformer Networks (STN) \cite{Jaderberg2015} as building blocks for these fMRI data processing neural networks. %% where external parameterisation is required.

In the current study, we use the aforementioned architecture to build a module able to adaptively select the most appropriate degree of smoothing for each volume in a data analysis task, allowing it to vary across time. This is a sharp contrast to previous work in the field, normally using preselected static smoothing or as a dynamic component within the GLM.
%(contrarily to most work in the filed which is either bound to the GLM or task independent, in the best case) allowing to vary the amount of smoothing across time. 
%It differs from \cite{Vilamala2016} in which the network goes beyond dealing with noise level by explicitly addressing the three main goals of the spatial smoothing step in fMRI in the design and training of the network. 
In contrast to previous work \cite{Vilamala2016}, we here explicitly consider the main goals of spatial smoothing while designing and training the network.
The provided solution can be used as a stand-alone module in regular fMRI pipelines %(\cite{Friston1995,Smith2004,Cox1996}) 
or as part of fMRI data processing neural networks.

The rest of the document is structured as follows: Section~\ref{sec:model} defines the architecture of the system and provides accurate implementation details, then we evaluate its performance for real fMRI brain data in Section~\ref{sec:evaluation}, where subjects were instructed to do simple finger tapping tasks. Finally, we provide some concluding remarks and future research directions.

%%%%%%% MODEL %%%%%%%%
\section{Model}
\label{sec:model}
Let $S = \left\{\mathbf{X}_n\right\}_{n=1}^N$ be a set containing fMRI brain volumes, each defined as $\mathbf{X} \in \mathbb{R}^{H \times W \times D}$ of height $H$, width $W$ and depth $D$. Our goal is to learn a function $\mathbf{Z} = s(\mathbf{X}, \sigma)$; $\mathbf{Z}$ being a volume of the same size as $\mathbf{X}$, corresponding to a spatially smoothed version of $\mathbf{X}$.

\begin{figure}[t]
\centering
\includegraphics[width=0.40\textwidth]{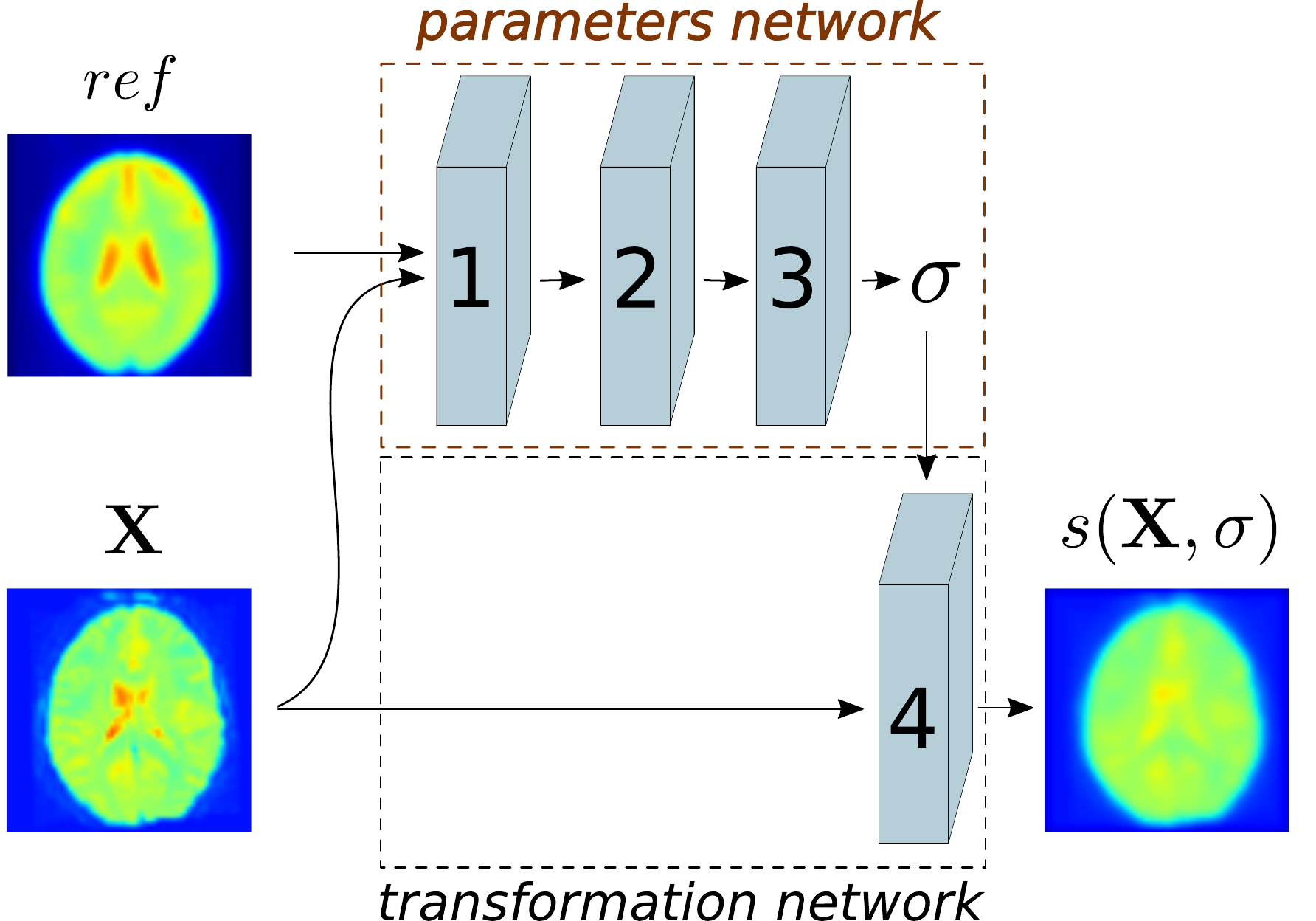}
\caption{Adaptive Smoothing neural network architecture.}
\label{fig:architecture}
\vspace{-5mm}
\end{figure}

Following the architecture proposed in \cite{Vilamala2016}, $s(\mathbf{X}, \sigma)$ is learnt by using a neural network made up of two subnetworks: \emph{the transformation network}, which smooths the input volumes by convolving them with a Gaussian filter, parameterised according to the values provided by \emph{the parameters network}, that calculates the most adequate filter size for the current volume. Next we describe the system, depicted in Fig.~\ref{fig:architecture}, in more detail.

\subsection{The transformation network}
Given an input volume and a value specifying the standard deviation parameter of the Gaussian filter ($\sigma$), first the appropriate filter is constructed and then it is convolved by the input. These two steps are more precisely defined in the following.

%------Gaussian filter construction
\subsubsection{Gaussian filter construction}
Consider a smoothing function as a continuous isotropic 3-dimensional Gaussian function with standard deviation $\sigma$:
\begin{equation}
g(x,y,z; \sigma) = \frac{1}{\left(\sqrt{2\pi} \sigma\right)^3} \exp\left\{-\frac{x^2 + y^2 + z^2}{2\sigma^2}\right\}.
\label{eq:gaussian}
\end{equation}

%\textcolor{red}{Derivative of the Gaussian formula wrt sigma. Is it required?}

%\textcolor{red}{Do we need to differentiate any of these grids formulas?}

A filter $\mathbf{Q}$ of varying size is obtained by sampling from Eq.~\ref{eq:gaussian} at specific locations specified by a grid $G$, defined as $G_i=(x_i, y_i, z_i)$, where $-\lfloor(t \cdot \sigma + 0.5)/2\rfloor \leq x_i, y_i,z_i \leq \lfloor(t \cdot \sigma + 0.5)/2\rfloor$; $t$ being the number of standard deviations from the mean where the Gaussian function is truncated. Finally, all values in the filter are renormalised to sum $1$.

Care must be taken for low $\sigma$ values. In particular, whenever $\sigma < 1.5/t$, the truncated discrete Gaussian function generates a single-cell $\mathbf{Q}$ filter, leading all partial derivatives of $\mathbf{Q}_{i,j,k}$ to be $0$; hence, causing the backpropagation gradient to disappear. This misbehaviour is alleviated by stochastically adding $1.0$ to $\sigma$ with probability $p$ (set to $0.1$ in all our experiments) for values below the mentioned threshold. Renormalising the single-cell $\mathbf{Q}$ filter must be done with care, since dividing by the sum of all elements cancels $\sigma$ out, causing the gradient to disappear. %Finally, possible division by $0$ artefacts in Eq.~\ref{eq:gaussian} must also be accounted for.
An alternative solution is to clip the values of $\sigma$ provided by the \emph{parameters network} above a certain threshold that avoids this problem altogether. %Moreover, if this threshold is set to be at least 2 times the size of the voxel resolution, RFT assumptions are met \cite{Worsley1995}, explicitly tackling goal $\#1$ of the spatial smoothing step (as enumerated in the Introduction). 
Moreover, it is typically reasonable to apply a certain minimum smoothing, in particular when RFT assumptions are sought to be fulfilled, where a smoothing of minimally $2$ times the voxel size is usually recommended \cite{Worsley1995}.   

%-----Volume smoothing
\subsubsection{Volume smoothing}
For a given convolutional smoothing operator $\mathbf{Q} \in \mathbb{R}^{H' \times W' \times D'}$, where $H' < H$, $W' < W$, $D' < D$, spatially smoothing a volume can be accomplished by sliding the operator over it, performing a convolution in each location. Finally, each element of the resulting smoothed volume $\mathbf{Z} \in \mathbb{R}^{H \times W \times D}$ is computed as:
\[
\mathbf{Z}_{h,w,d} = \sum_{i=0}^{H'}\sum_{j=0}^{W'}\sum_{k=0}^{D'} \mathbf{X}_{h+i,w+j,d+k} \cdot \mathbf{Q}_{i,j,k}.
\]

%\textcolor{red}{I've got the derivatives of the convolution wrt $\mathbf{X}$ and $\mathbf{Q}$ if required. But this is just a typical convolution}
%\textcolor{red}{Write the derivatives? Not to write them, but maybe just say that the whole thing is differentiable}

%------ The parameters network
\subsection{The parameters network}
So far, we have been omitting the origin of the important $\sigma$ parameter in the Gaussian filter, which is calculated using the \emph{parameters network}.
This network is implemented as a conventional feed forward neural network, with a final regressor outputting a $\sigma$ value for each volume. More precisely, the first layer in the network is a centring layer, where a provided reference volume is subtracted from all brain volumes. The rationale for using an external reference has multiple explanations. First, the resulting network learns to account for relative deviations of each volume to the reference (being it the average brain of all subjects involved in the study or a Montreal Neurological Institute (MNI) template, as appropriate for the study). Secondly, given the high dimensionality of the data in this domain it is very unlikely that all volumes in a study can be loaded in memory and processed at a time: an external volume eases the task of keeping a global reference. The second layer is a $2 \times 2 \times 2$ max-pooling module, with the purpose of quantifying the average noise level in the volume. By definition, after centring the data, the average voxel intensity in each volume lays roughly around $0$; a max-pooling layer only keeps the highest value in a neighbourhood, effectively increasing the average voxel intensity if noise is present in out-of-training sets. This has the effect of a multiplying factor to the value outputted by this network. The final layer is a fully-connected layer with a single output neuron with a softplus \cite{Dugas2001} activation function, ensuring non-negativity.

%----- Objective function
\subsection{Objective function}
In order to train the current module, we explicitly encode the goal of reducing inter-subject anatomical variability in the objective function. In particular, we seek to minimise this variability by smoothing individual volumes while penalising their deviation from the original volume. The final loss is computed as:
\begin{equation}
\footnotesize
L= \sum_i^N \Big[ s(\mathbf{X}_i, \sigma_i) - \frac{1}{N}\sum_n^N s(\mathbf{X}_n, \sigma_n)\Big]^2 + \lambda \Big[ \mathbf{X}_i - s(\mathbf{X}_i, \sigma_i)\Big]^2,
\label{eq:loss}
\end{equation}
$\lambda$ being a user-defined parameter controlling the trade-off between the smoothing level and deviation from original volumes (set to $0.5$ in all our experiments).

%----- Data augmentation
\subsection{Data augmentation}
\label{sec:data_augmentation}
To overcome the limitation of small sample size that are usually available in this domain and reduce overfitting due to the number of parameters in the network, we strongly rely on data augmentation techniques to obtain realistic samples \cite{Hauberg2016}. Specifically, given a volume from the training set $\mathbf{X}_i$, we generate a new sample volume $\mathbf{X}_a = \mathcal{T}_{\boldsymbol{\theta}}(\mathbf{X}_i)$;  where $\mathcal{T}_{\boldsymbol{\theta}}$ is a spatial transformation function. In particular, we use a neural network composed of two 3D STN, the first one able to apply affine transformations to the input volumes and the second one using Thin Plate Splines (TPS) \cite{Bookstein1989} to perform non-linear deformations (we use a sampling grid containing 192 free parameters as reference points). In contrast to their conventional use, the parameters of these modules are not calculated by their corresponding \emph{parameters networks}, but they are randomly sampled from a multivariate Gaussian distribution $\boldsymbol{\theta} \sim \mathcal{N}(\boldsymbol{\mu}, \boldsymbol{\Sigma})$,  
where $\boldsymbol{\mu}$ and $\boldsymbol{\Sigma}$ are calculated by aligning the first volume of each pair of subjects in the training set. Formally, let $\mathbf{X}_1^\alpha$ and $\mathbf{X}_1^\beta$ be the first volume in the series for subjects $\alpha$ and $\beta$, respectively. Our goal is to optimise
 $
\hat{\boldsymbol{\phi}}_{\alpha \beta} = \underset{\boldsymbol{\phi}}\argmin \left[\mathbf{X}_1^\alpha - \mathcal{T}_{\boldsymbol{\phi}}(\mathbf{X}_1^\beta)\right]^2
 $ 
for every pair of subjects, where $ \mathcal{T}_{\boldsymbol{\phi}}$ is the double 3D STN network defined in the previous paragraph. Then, $\boldsymbol{\mu}$ and $\boldsymbol{\Sigma}$ are set to the sample mean $\hat{\boldsymbol{\mu}}$ and covariance $\hat{\boldsymbol{\Sigma}}$ of matrix $\hat{\boldsymbol{\phi}}$.

%%%%%%% EMPIRICAL EVALUATION %%%%%%%%
\section{Empirical evaluation}
\label{sec:evaluation}
\subsection{Materials}
We use fMRI data \cite{Rasmussen2011} from a simple sequential finger tapping
paradigm, in which subjects alternated between 20 second blocks of
left and right finger tapping separated by 10 seconds of rest. Data
was recorded on a Siemens 3T scanner (Magnetom Trio) equipped with a
standard birdcage headcoil. Each of the 29 subjects' data consisted of
240 volumes with 3 mm isotropic resolution sampled at a repetition
time of 2.49 seconds. Further acquisition parameters can be found in
\cite{Rasmussen2011}. After basic preprocessing steps
including realignment and normalisation by standard settings in SPM12
(\url{http://www.fil.ion.ucl.ac.uk/spm/software/spm12}, revision 6685) each volume was labelled according to the left/right/rest condition starting from the second volume in each block.

%----- Artificially Perturbed Inputs
\subsection{Artificially perturbed inputs}

\subsubsection{Experimental setup}
Out of the $29$ subjects, the first $15$ were used as the training set and the remaining $14$ were set aside for testing. In this experiment, all  $240$ volumes per subject (normalised to $[0,1]$ using intra-subject extrema) were used irrespectively of their labelling. The fully-connected layer in the \emph{parameters network} was initialised using Xavier initialisation \cite{Glorot2010}. Reference volume was constructed by averaging the first volume of all subjects in the training set. At each epoch, new data was created by randomly selecting $7$ volumes from each subject in the training set and applying a random transformation to each one of them, as explained in Section~\ref{sec:data_augmentation}. The network was trained using Stochastic Gradient Descent (SGD) with Nesterov momentum for a number of internal epochs, after which new data was generated. All hyperparameters were selected using a Bayesian Optimisation (BO) strategy \cite{Snoek2012} validated on $7$ subjects from the test set, randomly picking $15$ volumes per subject and applying random transformations $\mathcal{T}_{\boldsymbol{\theta}}$. The whole procedure was repeated for the remaining $7$ subjects in the test set. Optimal results for both rounds were obtained by hyperparameters: epochs = $50$, internal epochs = $75$, learning rate = $0.7$ and momentum~=~$0.9$.

\subsubsection{Results}
\begin{figure}[t]
\vspace*{-0.1in}
\centering
\subfloat[White noise perturbation.]{\includegraphics[width=0.24\textwidth]{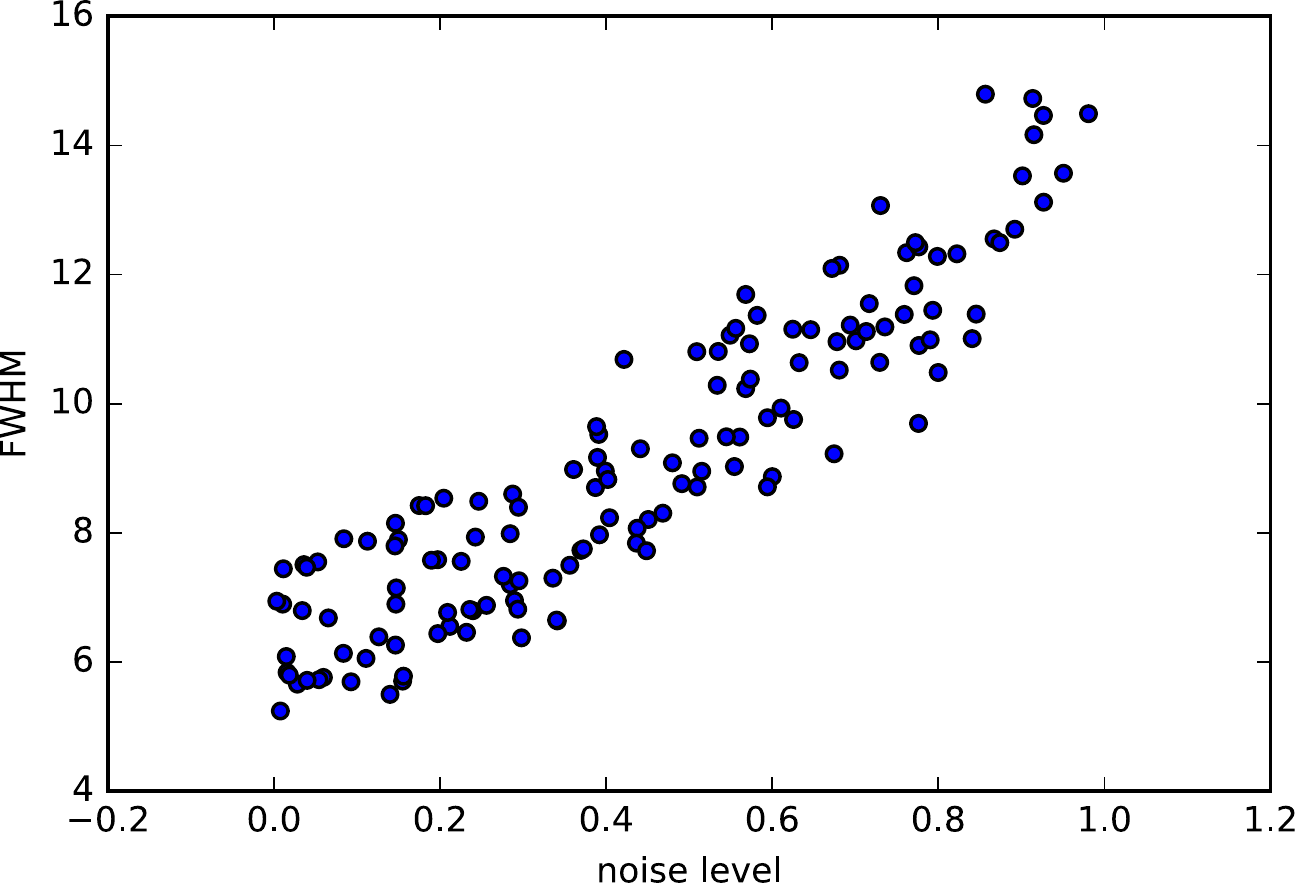}%
\label{fig:noise_fwhm}}
\hfil
\subfloat[Non-linear deformation.]{\includegraphics[width=0.24\textwidth]{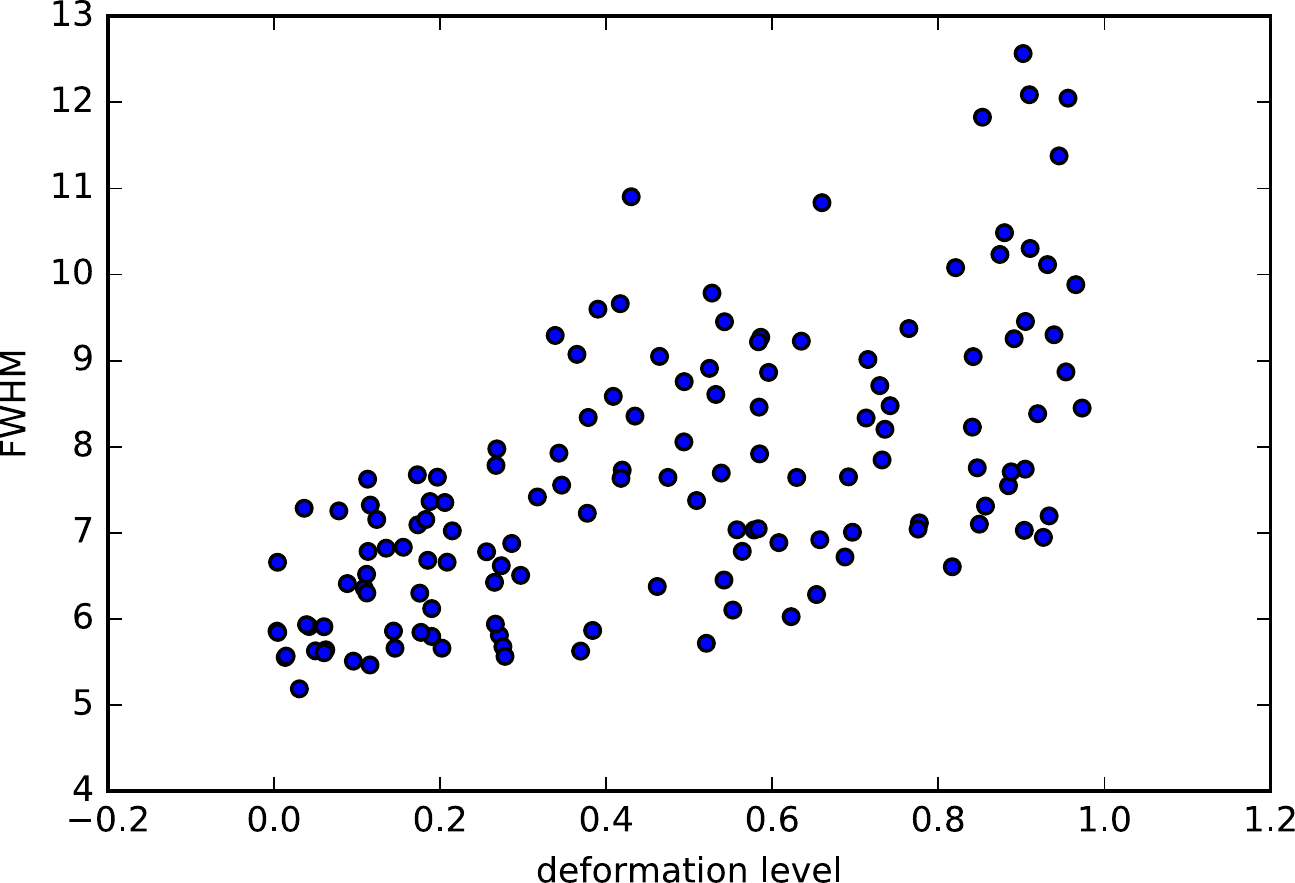}%
\label{fig:warp_fwhm}}
\caption{Correlation between the level of artificial perturbation applied to the volume ($\rho$) and FWHM (in mm.) proposed by the network.}
\vspace{-5mm}
\end{figure}

For assessing the capacity of our system to deal with noisy inputs, we randomly sample one volume from each of the $7$ remaining subjects in the test set and add uniform white noise $\sigma_n \sim \mathcal{U}(-a, a)$ to each voxel, where $a=m \cdot \rho$; $m=0.25$ being a predefined maximum noise level and $\rho \sim \mathcal{U}(0, 1)$ expressing the deformation percentage to be applied to each new volume. Reference volume is created by averaging the first volume of each subject in the set. This procedure is being repeated $10$ times, generating $2 \times 70$ new volumes with varying noise levels. These volumes are run through the network, which computes the corresponding filter size for each of the volumes. Results are converted to full width at half maximum (FWHM, a wide-spread measure of filter size). Fig.~\ref{fig:noise_fwhm} shows the correlation between the noise level $\rho$ and the FWHM in mm. 

%\begin{figure}[htb!]
%\centering
%\includegraphics[width=0.25\textwidth]{noise_fwhm}
%\caption{Correlation between the level of artificially added white noise ($\rho$) and FWHM (in mm.) proposed by the network.}
%\label{fig:noise_fwhm}
%\end{figure}
%
%\begin{figure}[htb!]
%\centering
%\includegraphics[width=0.25\textwidth]{warp_fwhm}
%\caption{Correlation between the level of deformation applied to the volume ($\rho$) and FWHM (in mm.) proposed by the network.}
%\label{fig:warp_fwhm}
%\end{figure}

Similarly, we use the previously introduced $\mathcal{T}_{\boldsymbol{\theta}}$ to generate $140$ brain volumes affected by different deformation levels. The maximum parameter levels are set to $0.05$ for translations, shearing and scaling, $\pi/32$ for rotations, and $0.05$ for non-linear warpings. Fig.~\ref{fig:warp_fwhm} depicts the correlation between the different deformation levels $\rho$ and the resulting FWHM in mm.

%----- Inter-subject anatomical differences
\subsection{Inter-subject anatomical differences}
In this experiment we want to evaluate the capacity of our pre-trained system to deal with the anatomical differences between subjects in the test set. We run our pre-trained network over all volumes in the test set, using the average of subjects' first volume as a reference. Table~\ref{tab:results} shows the average sum of squares difference between each smoothed subject's brain volume and the reference volume smoothed accordingly (i.e. first term of Eq.~\ref{eq:loss}). The penalty term is also calculated with the purpose to quantify the degree of loss incurred by each subject due to smoothing. Results show that when no smoothing is used (\emph{raw} column) the anatomical differences are highest, while when standard-practise $8$ mm. smoothing is applied, differences are diminished at the price of higher penalty to the raw image; adaptive smoothing presents a better trade-off between the two. \emph{Anat.} column in \emph{FWHM} shows the average smoothing the network calculates for each subject.

\begin{table}[t]
\setlength{\tabcolsep}{3.5pt}
\scriptsize 
  \caption{Inter-subject anatomical differences and assigned smoothing.}
  \label{tab:results}
  \centering
  \begin{tabular}{c | rrr | rr | rrr | cc}
   % \toprule
    &\multicolumn{3}{c | }{Differences} &\multicolumn{2}{c | }{Penalty} &\multicolumn{3}{c |}{FWHM} & \multicolumn{2}{c}{Acc. (in $\%$)} \\
    %\cmidrule{2-5}
    \# subj. & raw & 8mm & adap. & 8mm  & adap. & anat. & dec. & dec\_n & 8mm & adap.\\
        %\midrule
    16 & 16.1 &  9.3 & 10.8 & 18.0   & 12.8 & 5.8  & 3.7 &  9.8   & 58 &  58 \\
    17 & 21.5 &16.5 & 16.6 &  17.1  & 16.4 & 7.5  & 4.1 &  10.8 & 61 &  66\\
    18 & 15.3 &  7.7 & 9.5   &  17.8  & 12.6 & 5.7  & 3.3 &  9.7 & 62 &  56\\
    19 & 17.8 &  9.7 & 11.7 &  18.0  & 12.8 & 5.6  & 3.7 &  9.7 & 65 &  62\\
    20 & 19.5 &11.5 & 12.1 &  17.7  & 16.0 & 6.8  & 4.8 &  10.6 & 61 &  65 \\
    21 & 16.7 &11.2 & 12.3 &  16.4  & 12.3 & 6.0  & 5.0 &  10.5 & 58 &  63\\
    22 & 17.1 &  9.7 & 9.9 &  18.0  & 17.2 & 7.4  & 3.7 &  11.2 & 63 &  66\\
    23  & 19.1&13.2 & 13.4 &  16.8  & 16.1 &7.4   & 3.7 &  11.2 & 55 &  56\\
    24  & 20.5&16.4 & 17.3 &  15.9  & 10.9 & 5.3  & 5.1 &  10.1 & 56 &  62\\
    25  & 16.1&  9.6 & 10.1 &  17.8  & 15.8 & 6.7  & 3.1 &  11.1  & 53 &  62\\
    26  & 17.0&11.2 & 12.4 & 17.1   & 12.4 & 5.8  & 4.5 &  10.4  & 59 &  65\\
    27  & 19.7& 13.3& 13.8 &  18.3  & 16.3 & 6.7  & 2.6 &  10.7  & 59 &  64\\
    28  & 18.2&10.1 & 10.6 &  17.6  & 16.3 & 7.1  & 3.5 &  11.3 & 62 &  62\\
    29  & 15.4&  8.7 & 10.4 &  16.5  & 11.5 & 5.6  & 3.3 &  10.3  & 61 &  60\\
    Avg.  & 17.9 &  11.3 & 12.2 & 17.4 & 14.2 & 6.4 & 3.9 & 10.5 & 59 &  62

  \end{tabular}
  \\
 \vspace{2mm}
{\justifying \noindent \emph{Differences} show the average sum of squares distance between each subject and reference volume for a variety of smoothings: raw data, fixed 8mm. and using our adaptive method. Similarly, \emph{Penalty} expresses the same measure between smoothed and raw instances. \emph{FWHM} shows the kernel width selected by our adaptive method for different analytical tasks: inter-subject anatomical differences, brain decoding and decoding with noise. \emph{Acc.} column reports the obtained accuracies for the noisy brain decoding task.\par}
     \vspace{-5mm}
\end{table}

\begin{figure}[t]
\centering
\includegraphics[width=0.26\textwidth]{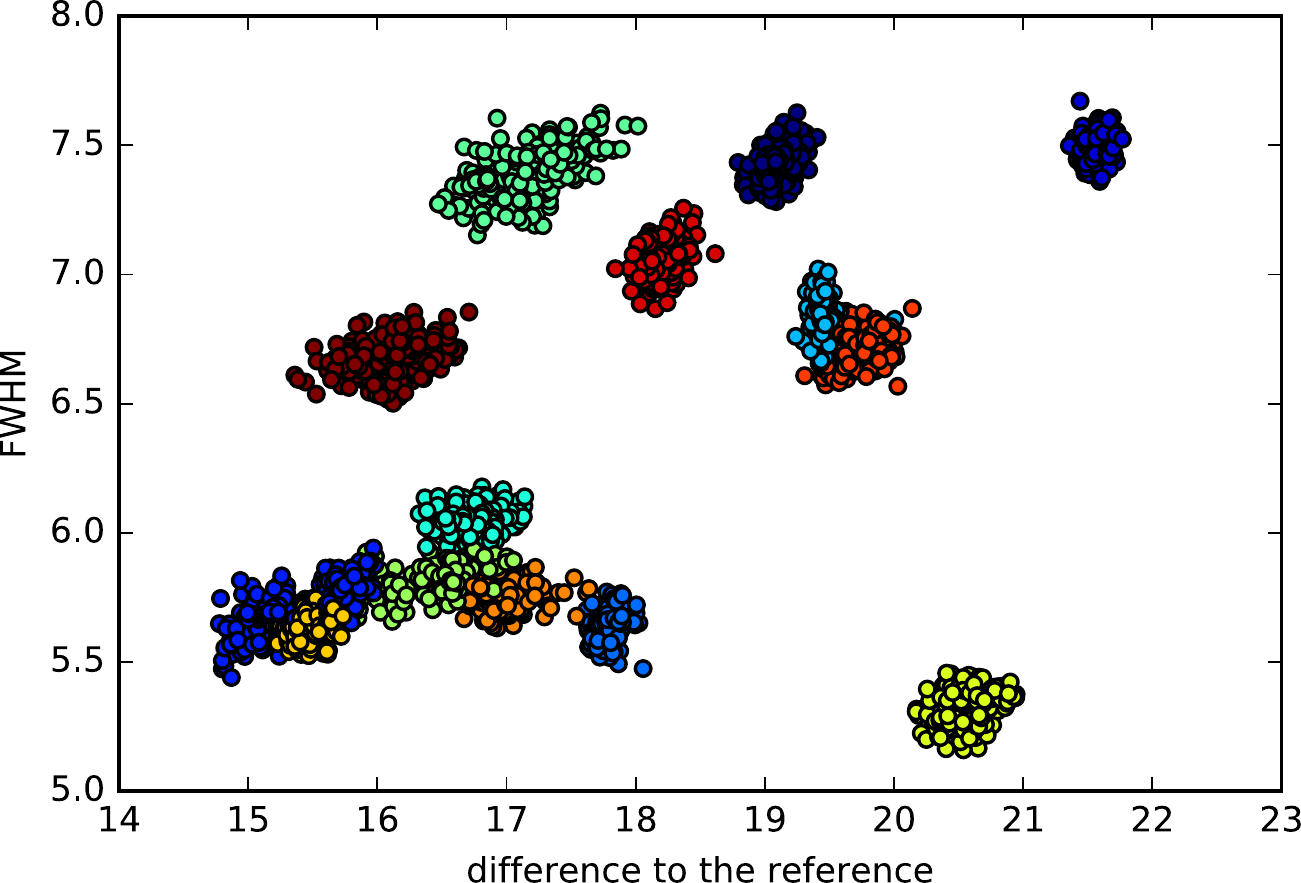}
\caption{Correlation between volume difference to the reference and FWHM (in mm.) proposed by the network. Each colour represents a different subject.}
\label{fig:diff_fwhm}
\vspace{-5mm}
\end{figure}

Looking at Fig.~\ref{fig:diff_fwhm}, the system shows general consistency in assigning varying smoothing levels to volumes according to their distance to the reference volume. Likewise, it is interesting to notice that although the use of different smoothing accounting for inter-subject dissimilarities seems plausible, intra-subject differences are quite minimal, therefore, there is no real need to use a per-volume smoothing at this stage. 

%----- Brain decoding task
\subsection{Brain decoding task}
The strongest point of the proposed adaptive smoothing is its capacity to adjust the degree of smoothing to each specific data analysis task, as part of a multilayer neural network. In this section, we add a decoding module after the smoothing one aiming at classifying each brain volume as the state where the subject was finger tapping her right or left hand; hence, brain volumes labelled as rest were discarded, using only $200$ volumes per subject. The decoding module is composed of a fully-connected layer with single linear output, to which a variant of batch-normalisation \cite{Ioffe2015} is applied before a sigmoid activation. Importantly, all $200$ volumes of the same subject conform the mini-batch, which coupled to the offline nature of the experiments, allows for using current batch statistics, being it training, validation or testing. All settings and training procedures were defined as in the previous experiment (i.e., 15 subjects for training, 7 for validation and 7 for test), but using the pre-trained weights as a starting point for the \emph{parameters network}. No data augmentation was used and binary cross-entropy between predicted values and targets was the loss function of choice. Network was trained, allowing fine-tune of the \emph{parameters network}, using SGD with hyperparameters set to be epochs = $100$, learning rate = $0.01$.

Here, we allow the adaptive smoothing to freely choose the most adequate smoothing (i.e., adapting weights in \emph{the parameters network}) providing the highest classification accuracy. Notice that we are not aiming at improving accuracy in this setting (due to the powerful motor activation paradigm, the classification can be obtained almost perfectly without smoothing), but we rather want to observe the behaviour of the adaptive smoothing module. As expected, \emph{dec.} column in Table~\ref{tab:results} shows the average smoothing parameter for each subject, which tends to vanish for the current task. In order to challenge the network, Gaussian white noise with $\sigma=0.25$ is artificially added to the input volumes, forcing the smoothing kernel to be widened: \emph{dec\_n} column shows the most appropriate average smoothing for each subject, and last two columns present the corresponding classification accuracies. Although there seems to be a slight improvement in classification accuracy when our method is employed, further experiments should be done in datasets encoding tasks highly dependent on the smoothing step to appreciate the extent of our method.
 
\section{Conclusion}
Using a novel implemented spatial smoothing module, we here demonstrate that it is possible to construct differentiable modules of neural networks that form flexible and dynamic analysis pipelines for fMRI data. As a proof of concept we additionally show that it is feasible to train such networks end-to-end, which is an important step towards truly optimal processing pipelines. Future work will focus on the development of a standard fMRI pipeline in form of a neural network, including determining spatial location activations.

%The use of fMRI data processing neural networks requires the development of differentiable modules, which assembled together, provide the desired outcome. We have proposed a module able to spatially smooth brain volumes at varying levels over time adapted to each specific analysis task. 


\section*{Acknowledgment}
\footnotesize
{
This project has received funding from the European Union's Horizon 2020 research and innovation programme under the Marie Sklodowska-Curie grant agreement No 659860. We gratefully acknowledge the support of NVIDIA Corporation with the donation of the GPUs used for this research.
}



\bibliographystyle{IEEEtran}
\bibliography{references}
%\begin{thebibliography}{1}
%\end{thebibliography}

\end{document}